\documentclass[letterpaper]{article} 
\usepackage{aaai2026}  
\usepackage{times}  
\usepackage{helvet}  
\usepackage{courier}  
\usepackage[hyphens]{url}  
\usepackage{graphicx} 
\usepackage{natbib}  
\usepackage{caption} 
\usepackage{algorithm}
\usepackage{algorithmic}

\usepackage{newfloat}
\usepackage{listings}

\DeclareCaptionStyle{ruled}{labelfont=normalfont,labelsep=colon,strut=off} 

\definecolor{myrowcolor}{HTML}{F2F2F2}
\definecolor{keywordcolor}{HTML}{770489}

\lstdefinelanguage{SPARQL}{
  keywords={SELECT, WHERE, FILTER, OPTIONAL, PREFIX, LIMIT, DISTINCT, GROUP, ORDER, BY, DESC, COUNT},
  sensitive=true,
  morecomment=[l]{\#},
}

\floatstyle{ruled}
\newfloat{listing}{tb}{lst}{}
\floatname{listing}{Listing}
\usepackage{booktabs}
\usepackage{tabularx}
\usepackage[table]{xcolor}

\newcommand{\ourkb}{\mbox{GPTKB v1.5}}

\title{\ourkb{}: A Massive Knowledge Base for Exploring Factual LLM Knowledge}

\author{
    Yujia Hu\textsuperscript{\rm 1},
    Tuan-Phong Nguyen\textsuperscript{\rm 2},
    Shrestha Ghosh\textsuperscript{\rm 3},
    Moritz Müller\textsuperscript{\rm 1},
    Simon Razniewski\textsuperscript{\rm 1}
}

\affiliations{
    \textsuperscript{\rm 1}ScaDS.AI Dresden/Leipzig \& TU Dresden, Germany\\
    \textsuperscript{\rm 2}VNU University of Engineering and Technology, Hanoi, Vietnam\\
    \textsuperscript{\rm 3}University of Tübingen, Germany\\
    yujia.hu@tu-dresden.de,
    tuanphong@vnu.edu.vn,
    shrestha.ghosh@uni-tuebingen.de,
    simon.razniewski@tu-dresden.de
}

\begin{document}

\maketitle

\begin{abstract}
Language models are powerful artifacts, yet their factual knowledge is still poorly understood, and inaccessible to ad-hoc browsing and scalable statistical analysis. This demonstration introduces \ourkb, a densely interlinked 100-million-triple knowledge base (KB) built for \$14,000 from GPT-4.1, using the GPTKB\linebreak methodology for massive-recursive LLM knowledge materialization.
This demo focuses on three use cases: (1) link-traversal-based LLM knowledge exploration, (2) SPARQL-based structured LLM knowledge querying, (3) comparative exploration of the strengths and weaknesses of LLM knowledge.
Massive-recursive LLM knowledge materialization is a groundbreaking opportunity both for the systematic analysis of LLM knowledge, as well as for automated KB construction. 
\end{abstract}

\begin{links}
    \link{Website}{https://gptkb.org/}
\end{links}

\section{Introduction}
 
Large Language Models (LLMs) have demonstrated the ability to store a surprising amount of factual knowledge with remarkable accuracy \cite{petroni2019language}. However, the scope and nature of this factual knowledge are still poorly understood. Several works have sought to benchmark the factual knowledge encoded in LLMs~\cite{jiang2020how,roberts2020how,wang2021can,sun2023headtotail,veseli2023evaluating}. Yet, these efforts are inherently limited by their reliance on pre-selected samples, introducing an availability bias \cite{kahnemann}: they only assess knowledge that is already anticipated by the experimenters. As a result, factual knowledge (or beliefs) not foreseen by the study design goes unnoticed.

\begin{figure}[t]
    \centering
    \includegraphics[width=.75\columnwidth]{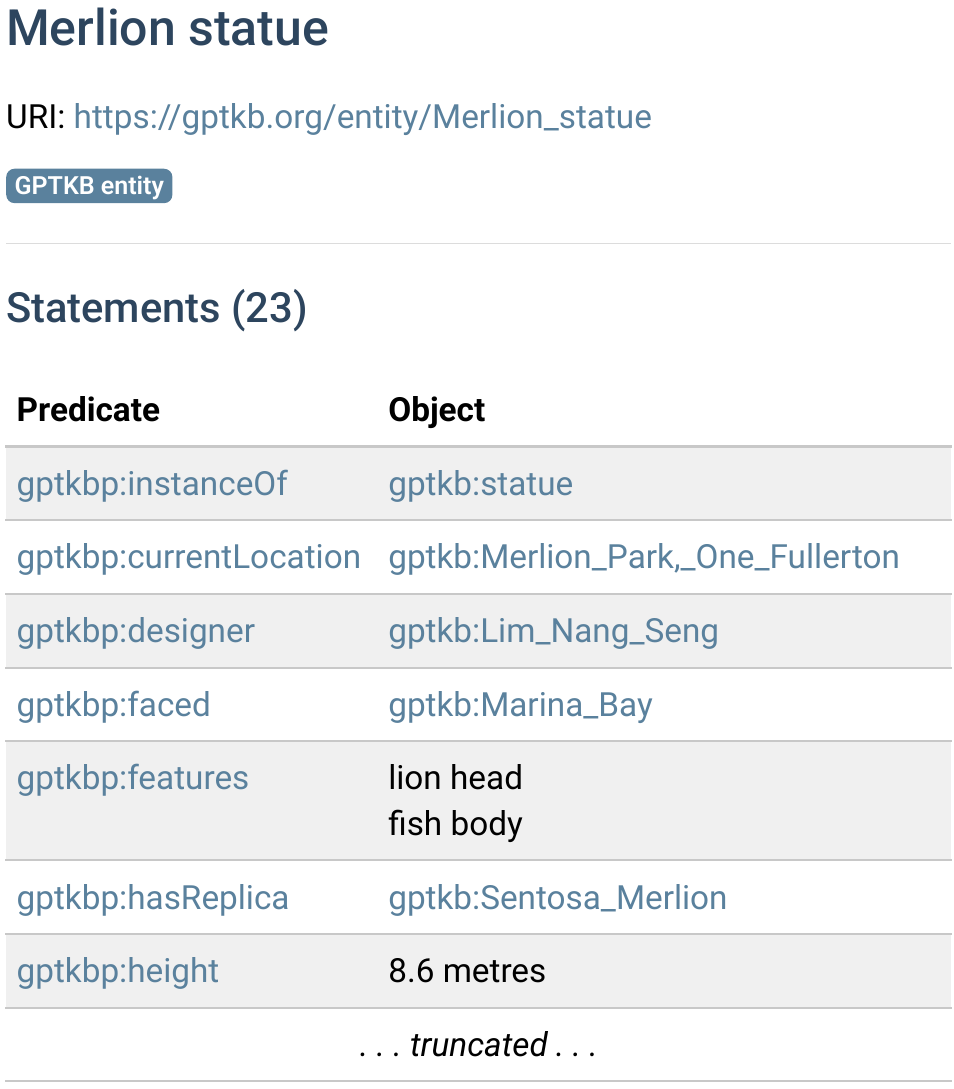}
    \caption{GPTKB page for \textit{Merlion statue}.}
    \label{fig:merlion}
\end{figure}

In \cite{hu2024gptkbcomprehensivelymaterializingfactual}, we proposed the GPTKB methodology for recursively extracting and materializing factual LLM knowledge at massive scale. This recursive process enables us to overcome the availability bias inherent in sample-based evaluations. In this demo we present \ourkb{}, a 100M-triple KB entirely built from GPT-4.1, and show how one can explore and traverse the KB, query it, and compare LLMs by their knowledge. A screenshot from our web interface is shown in Figure~\ref{fig:merlion}.

\section{GPTKB Methodology}\label{sec:methodology}

The GPTKB methodology \cite{hu2024gptkbcomprehensivelymaterializingfactual} combines a recursive knowledge elicitation process with a post-hoc knowledge consolidation phase.\\
\textbf{Knowledge elicitation.}
Starting from a seed entity, the LLM is prompted to return knowledge about it in the form of triples. New named entities in these triple objects are identified via LLM-based named-entity recognition (NER) and are enqueued for further elicitation in a recursive BFS-based graph exploration process. Constrained decoding is used to make sure that outputs stay within the triple format.\\
\textbf{Knowledge consolidation.}
To address the redundancy and variance introduced during knowledge elicitation, post-hoc knowledge consolidation is performed. In particular, we apply a greedy clustering algorithm to iteratively merge relations and classes into more frequent ones, given a sufficiently high label embedding similarity.

\section{\ourkb{} Construction}

As the basis for this demo, we executed the GPTKB methodology \cite{hu2024gptkbcomprehensivelymaterializingfactual} using the GPT-4.1 LLM, one of the strongest frontier models available in summer 2025. Following the paradigm in Section~\ref{sec:methodology}, we extracted knowledge starting with the seed entity \textit{Vannevar Bush}, for a total BFS depth of 10 layers. The whole process cost \$14,136 for OpenAI API calls and took 18 days. The final KB contains 100 million triples derived from 6.1 million entities in total, organized into 381k relations and 32k classes. 
We provide statistics of \ourkb{} in Table~\ref{tab:statistics}. To facilitate data interchange, we also converted GPTKB into RDF format, and provide it as a 4 GB download.

We performed two \textbf{quality evaluations}. An automated method based on web search, like in \cite{hu2024gptkbcomprehensivelymaterializingfactual}, using 1,000 random triples, and a manual assessment of 100 triples. Both annotations agree in the fraction of correct triples (75.5\% and 75\%), while the automated evaluation reported a slightly higher degree of incorrect ones (19.5\% versus 14\% in manual). In both cases, the truth of some triples remains undecidable, mostly, because parts of them are semantically incomprehensible.

\begin{table}[t]
\centering
\resizebox{\columnwidth}{!}{%
\small
\begin{tabular}{@{}ll@{}}
\toprule
\textbf{Entities} & 6.1M \\
\textbf{Triples} & 100M (120M with meta-relations) \\
\textbf{Relations} & 936k (381k after canonicalization) \\
\textbf{Classes} & 220k (32k after canonicalization) \\
\textbf{Triple objects} & 59M entities, 41M literals \\
\textbf{Avg. triples/entity} & 16.3 \\
\textbf{Avg. label length} & 19.8 characters \\
\textbf{Subject-precision} & 85.3\% Verifiable, 3.4\% Plausible\\
 & 11.3\% Unverifiable \\
\textbf{Subjects in Wikidata} & 43\% \\
\textbf{Triple-precision} & 75.5\% True, 5.0\% Plausible, 19.5\% False \\ 
\textbf{Cost of API-calls} & \$14,136\\
\bottomrule
\end{tabular}
}
\caption{Statistics of \ourkb{}.}
\label{tab:statistics}
\end{table}

\section{Demonstration Experience}
We host our demo on an interactive web interface. The demonstration experience is divided into three parts: (1) link-based graph exploration, (2) structured SPARQL queries, (3) comparative analysis.

\noindent
\textbf{Link-based Knowledge Graph Exploration.}
Data about specific entities can be accessed directly from the start page (\url{https://gptkb.org}), via a search field (top-right corner), or directly by URL (\url{https://gptkb.org/entity/<NAME>}). For example, in Figure~\ref{fig:merlion}, from \textit{Merlion statue}, one can navigate onwards to \textit{Lim Nang Seng} or \textit{Marina Bay}. Besides the LLM-based core data, we also provide two post-hoc added meta-relations, \textit{bfsLayer} and \textit{bfsParent}, which allow to locate an entity in the graph. This way, one can interactively explore the resulting knowledge graph. 

\noindent
\textbf{SPARQL Querying to Understand LLM Knowledge and Gaps.}
A core intention of GPTKB is to enable LLM factual knowledge analysis at scale. While traditional analyses (e.g., \cite{kotek2023gender}) rely on small-scale repetitive prompting or problem-specific templates, the materialized knowledge in GPTKB enables large-scale analysis at the fingertip of modern database technology. For this purpose, the GPTKB content is stored in a Virtuoso Triple store and made accessable through a SPARQL query interface at \url{https://gptkb.org/query/}. For example, just what kind of entities does GPT know about? An overview is provided by:\\[5pt]
\begin{lstlisting}
SELECT ?o (COUNT(*) AS ?ofreq)
WHERE { ?s gptkbp:instanceOf ?o. } 
GROUP BY ?o ORDER BY DESC(?ofreq) LIMIT 100
\end{lstlisting}
\noindent
{\renewcommand{\arraystretch}{1.1}\footnotesize\ttfamily
\rowcolors{2}{white}{myrowcolor}
\begin{tabularx}{\columnwidth}{Xl}
\textbf{o} & \textbf{ofreq} \\
gptkb:person & 1,077,803 \\
gptkb:human & 138,646 \\
gptkb:film & 120,497 \\
gptkb:company & 118,993 \\
gptkb:book & 111,414 \\
gptkb:song & 103,538 \\
gptkb:fictional\_character & 90,499 \\
\end{tabularx}\medskip}

Similarly, lack of symmetry is a long-standing problem of LLM knowledge representation~\cite{reversalcurse}, but how prevalent is this in the long tail of a frontier model? The following query asks for the fraction of spousal triples that are present in both directions.\\[5pt]
\begin{lstlisting}
SELECT (COUNT(?a_both) AS ?numMutual) (COUNT(?a) AS ?total) ((COUNT(?a_both) * 1.0) / COUNT(?a) AS ?fraction)
WHERE {{ SELECT DISTINCT ?a ?b 
            WHERE { ?a gptkbp:spouse ?b.}}
        OPTIONAL { ?b gptkbp:spouse ?a. BIND(?a AS ?a_both) }}
\end{lstlisting}
\noindent
{\renewcommand{\arraystretch}{1.1}\footnotesize\ttfamily
\rowcolors{2}{white}{myrowcolor}
\begin{tabularx}{\columnwidth}{Xll}
\textbf{numMutual} & \textbf{total} & \textbf{fraction}\\
65,339 & 402,333 & 0.162 \\
\end{tabularx}\medskip}
As we can see here, asymmetry is also dominant in GPT-4.1's factual knowledge.

\noindent
\textbf{Comparing LLM Results.}
The ``Compare'' menu item allows to contrast the factual knowledge of different LLMs side-by-side. We provide results for a total of five LLMs: two Llama variants (3.3-70B-Instruct, 4-Scout-17B-16E-Instruct), two GPT variants (4o-mini and 4.1), and DeepSeek-R1 as dedicated reasoning model. For each of these models, we provide the outputs on a selected diverse set of 100 entities. We used the same prompt as in \cite{hu2024gptkbcomprehensivelymaterializingfactual} for all models.

Once two models and an entity are selected, the page shows the returned triples side-by-side, along with flags denoting the correctness of each statement. The correctness is computed using an automated RAG-based web validation framework, as in \cite{hu2024gptkbcomprehensivelymaterializingfactual}. Furthermore, the top of the page presents statistics regarding totals.

For example, for Ilya Sutskever, one can see that GPT produces significantly more triples than Llama (40 versus 14), and most triples of both LLMs could be web-verified. GPT produces the more readable \textit{instanceOf} values, while Llama here uses Wikidata's Q5-identifier for humans. Notably, both models assert a wrong birth date (w.r.t. Wikipedia). This way, one can get insights into the capabilities as well as failures of LLM knowledge.

\bibliography{references}

\end{document}